\documentclass{article}
\usepackage{graphicx} %

\author{
  \textbf{Lennart Bramlage}\textsuperscript{\textrm{1,2,*}} \quad
  \textbf{Cristóbal Curio}\textsuperscript{\textrm{1,2}} \\
  \textsuperscript{1}Reutlingen University, \textsuperscript{2}University of Tübingen \\
  \textsuperscript{\textrm{*}}Corresponding author: lennart.bramlage@reutlingen-university.de
}

\usepackage[preprint]{neurips_2025}

\usepackage[utf8]{inputenc} %
\usepackage[T1]{fontenc}    %
\usepackage{hyperref}       %
\usepackage{url}            %
\usepackage{booktabs}       %
\usepackage{amsfonts}       %
\usepackage{nicefrac}       %
\usepackage{microtype}      %
\usepackage[dvipsnames]{xcolor}         %
\usepackage{amsmath}
\usepackage[english]{babel}
\usepackage{natbib}
\usepackage[ruled,vlined]{algorithm2e}
\usepackage{float}
\usepackage{makecell}
\usepackage{wrapfig}
\usepackage{subcaption}
\usepackage{enumitem}
\newcommand*{\fullref}[1]{\hyperref[{#1}]{\autoref*{#1} \nameref*{#1}}} %
\DeclareMathOperator*{\argmin}{arg\,min}
\newtheorem{proposition}{Proposition}

\newtheorem{corollary}{Corollary}

\title{Principled Input-Output-Conditioned Post-Hoc Uncertainty Estimation for Regression Networks}

\begin{document}

\maketitle

\begin{abstract}
    Uncertainty quantification is critical in safety-sensitive applications but is often omitted from off-the-shelf neural networks due to adverse effects on predictive performance. Retrofitting uncertainty estimates post-hoc typically requires access to model parameters or gradients, limiting feasibility in practice. We propose a theoretically grounded framework for post-hoc uncertainty estimation in regression tasks by fitting an auxiliary model to both original inputs and frozen model outputs. Drawing from principles of maximum likelihood estimation and sequential parameter fitting, we formalize an exact post-hoc optimization objective that recovers the canonical MLE of Gaussian parameters, without requiring sampling or approximation at inference. While prior work has used model outputs to estimate uncertainty, we explicitly characterize the conditions under which this is valid and demonstrate the extent to which structured outputs can support quasi-epistemic inference. We find that using diverse auxiliary data, such as augmented subsets of the original training data, significantly enhances OOD detection and metric performance. Our hypothesis that frozen model outputs contain generalizable latent information about model error and predictive uncertainty is tested and confirmed. Finally, we ensure that our method maintains proper estimation of input-dependent uncertainty without relying exclusively on base model forecasts. These findings are demonstrated in toy problems and adapted to both UCI and depth regression benchmarks. Code: \url{https://github.com/biggzlar/IO-CUE}.
\end{abstract}

\section{Introduction}

Uncertainty quantification (UQ) is critical for deploying machine learning systems in safety-sensitive contexts \citep{ayhan2018test,begoli2019need}, yet it remains underutilized in practice due to its negative impact on model performance \citep{seitzer2022pitfalls}. Retrofitting existing models with UQ capabilities often requires costly retraining or access to model parameters or intermediate representations -- conditions frequently unmet in real-world scenarios.

We propose a scalable, post-hoc framework for UQ based on sequential distribution parameter fitting \citep{cawley2004heteroscedastic,yuan2004doubly}, which circumvents these limitations. Unlike MCMC \citep{neal2011mcmc} or Laplace-based methods \citep{ritter2018scalable}, our approach does not require expensive test-time sampling or access to model internals. It also only requires a fraction of the data required for training from scratch and does not impact base model performance negatively. To our knowledge, this approach has not been rigorously formalized in prior work.

In addition, we examine how frozen model predictions can reveal latent structure indicative of epistemic uncertainty, enabling effective out-of-distribution (OOD) detection and uncertainty estimation. We provide both theoretical justification and empirical evidence across regression and vision-based tasks for these claims.

\subsection{Contributions}
\begin{itemize}
    \item We introduce a theoretically grounded and \textbf{Input-Output-Conditioned Uncertainty Estimation (IO-CUE)} framework based on sequential disjoint parameter estimation, applicable to frozen base models post-hoc without requiring access to parameters or internal states.
    
    \item We provide a detailed analysis of how structured model outputs can implicitly encode epistemic uncertainty, enabling post-hoc estimators to approximate OOD sensitivity under black-box constraints.

    \item We propose a principled approach to enhancing post-hoc model characterization through targeted data augmentations, significantly improving uncertainty estimation and OOD detection across both simple and complex regression domains.
\end{itemize}

\section{Related Work}
\textbf{Uncertainty quantification.} A wide range of methods exist for UQ, but we focus on gradient-based optimization of the negative log-likelihood (NLL) objective for heteroscedastic regression \citep{nix1994estimating}. Under a Gaussian assumption, this corresponds to learning both the predictive mean and input-dependent variance, where the latter reflects aleatoric uncertainty \citep{der2009aleatory}. Epistemic uncertainty, which arises from model misspecification \citep{lahlou2021deup} or parameter uncertainty \citep{blundell2015weight}, is typically estimated via ensemble disagreement, MC dropout, or other sampling-based approaches \citep{kendall2017uncertainties,gal2016dropout,lakshminarayanan2017simple}. More recent methods attempt to jointly estimate both uncertainty types through higher-order distributional objectives \citep{sensoy2018evidential,amini2020deep,sale2023second,bramlage2023plausible}, though this has been criticized as ill-posed since epistemic uncertainty, by definition, cannot be learned from finite data alone \citep{bengs2022pitfalls,meinert2023unreasonable,jurgens2024epistemic}.

\textbf{Sequential parameter estimation.} In classical settings, maximum likelihood estimation (MLE) of heteroscedastic Gaussian models proceeds sequentially: the mean is first estimated via least squares, yielding the MAP estimate under a Gaussian prior, after which the variance can be estimated conditioned on this fixed mean \citep{cawley2004heteroscedastic,yuan2004doubly}. This separation is statistically sound and exploits the conditional structure of the likelihood. However, modern approaches often optimize the negative log-likelihood (NLL) jointly over mean and variance parameters via gradient descent. This poses two challenges: first, the Gaussian NLL is not jointly convex, only conditionally convex in each parameter, leading to unstable or suboptimal convergence; second, when one of the parameters is far from its MAP estimate, the other is estimated under misspecified conditions, further compounding error. To address these issues, prior work has explored reparameterizing the Gaussian using its natural parameters \citep{le2005heteroscedastic,immer2023identifiability}, applying regularization to stabilize training \citep{seitzer2022pitfalls}, or using alternating optimization schemes that decouple updates \citep{detlefsen2019reliable}.

\textbf{Post-hoc UQ.} Given the above challenges, post-hoc UQ offers a practical alternative. Estimating the mean via least-squares yields the MAP estimate, after which variance parameters can be fit more effectively. Several methods build on this by sampling from the posterior over weights using approximate gradient-based updates \citep{maddox2019simple}, MCMC \citep{neal2011mcmc,betancourt2017conceptual,chen2014stochastic}, Dropout injection \citep{ledda2023dropout} or Laplace approximations \citep{mackay1992practical,ritter2018scalable,daxberger2021laplace}. However, these require access to model parameters and multiple forward passes through sampled networks at inference-time. In contrast, \citet{upadhyay2022bayescap} propose estimating uncertainty directly from base model predictions, enabling black-box applicability and single-forward pass inference. Their approach motivates our more rigorous formulation of the necessary conditions for proper post-hoc UQ, as well as the so-far unexplored capacity of structured output spaces to encode epistemic model uncertainty.

\section{Methodology}
\subsection{Problem Statement}
We consider the general regression problem of mapping inputs $x \in \mathbb{R}^D$ to outputs $y \in \mathbb{R}^K$, where both spaces may be of arbitrary dimensionality. We denote a base model $f: \mathcal{X} \mapsto \mathbb{R}^K$ with parameters $\theta$ and predictions $\mu_{\theta}(x) = \hat y = f(x; \theta)$ without the capability to infer uncertainty. Assuming a frozen base model and no access to intermediate representations, we want to train an auxiliary model $g: \mathcal{X} \times \mathcal{Y} \mapsto \mathbb{R}^+$ with parameters $\phi$ on a probe dataset $\mathcal{D}_{probe}$ to estimate the uncertainty of the base model's predictions $\sigma_{\phi}(x, y) = g(x, f(x); \phi)$. \emph{We aim to (1) capture data uncertainty in-distribution (ID), and (2) provide relative epistemic uncertainty forecasts for, e.g., OOD detection.}

\subsection{Preliminaries}
\textbf{Sources of uncertainty.} For the sake of simplicity, we assume additive Gaussian noise in the following set of definitions. Heteroscedastic aleatoric uncertainty arises from noise in the data and is typically represented as $\sigma^2(x) = \mathrm{Var}(y \vert x)$, such that $y \sim \mathcal{N}(\mu(x), \sigma^2(x))$, where $\mu(x)$ and $\sigma^2(x)$ represent the ground-truth functions for mean and variance, respectively. Epistemic uncertainty in neural networks is generally described as uncertainty in the model's parameters. In practice, it is often represented by evaluating the variance of the predictions of several models drawn from the posterior distribution of parameters, given the data $\mathbb{V}_{\theta \sim p(\theta \mid \mathcal{D})}[f(x; \theta)]$ \citep{gal2016dropout,d2021repulsive,abe2022deep}. As such, the total uncertainty \citep{depeweg2018decomposition, wimmer2023quantifying} is given by
\begin{equation}
    \mathrm{Var}(y \vert x, \mathcal{D}) = \sigma^2(x) + \mathbb{V}_{\theta}\left[\mu_{\theta}(x)\right].
\end{equation}

\textbf{Estimating uncertainty.} While a multitude of approaches exist, we focus on gradient-based maximum likelihood estimation (MLE) of the mean and variance functions $\mu_{\theta}(x)$ and $\sigma_{\theta}^2(x)$. Such models are trained concurrently with an appropriate objective function like the Gaussian NLL
\begin{equation}
    \mathcal{L}(x; \theta) = \frac{(y_i - \mu_{\theta}(x))^2}{2 \sigma_{\theta}^2(x)} + \frac{1}{2}\log\sigma_{\theta}^2(x) + C.
\end{equation}
As a proper scoring rule \citep{gneiting2007strictly}, the NLL is minimized if and only if $\mu_{\theta}(x) = \mu(x)$ and $\sigma_{\theta}^2(x) = \sigma^2(x)$, i.e., $\sigma_{\theta}^2(x)$ converges to the true aleatoric variance $\sigma^2(x)$. Epistemic uncertainty is not captured by the NLL and is thus commonly approximated by a discrete ensemble \citep{lakshminarayanan2017simple} or multiple Monte-Carlo Dropout passes \citep{gal2016dropout,kendall2017uncertainties, folgoc2021mc}, such that $\mathbb{V}_{\theta}\left[\mu_{\theta}(x)\right] \approx \frac{1}{M-1} \sum_{m=1}^M \left( \mu_{m}(x) - \bar \mu(x) \right)^2$.

\begin{figure}[t]
    \centering
    \includegraphics[width=\textwidth]{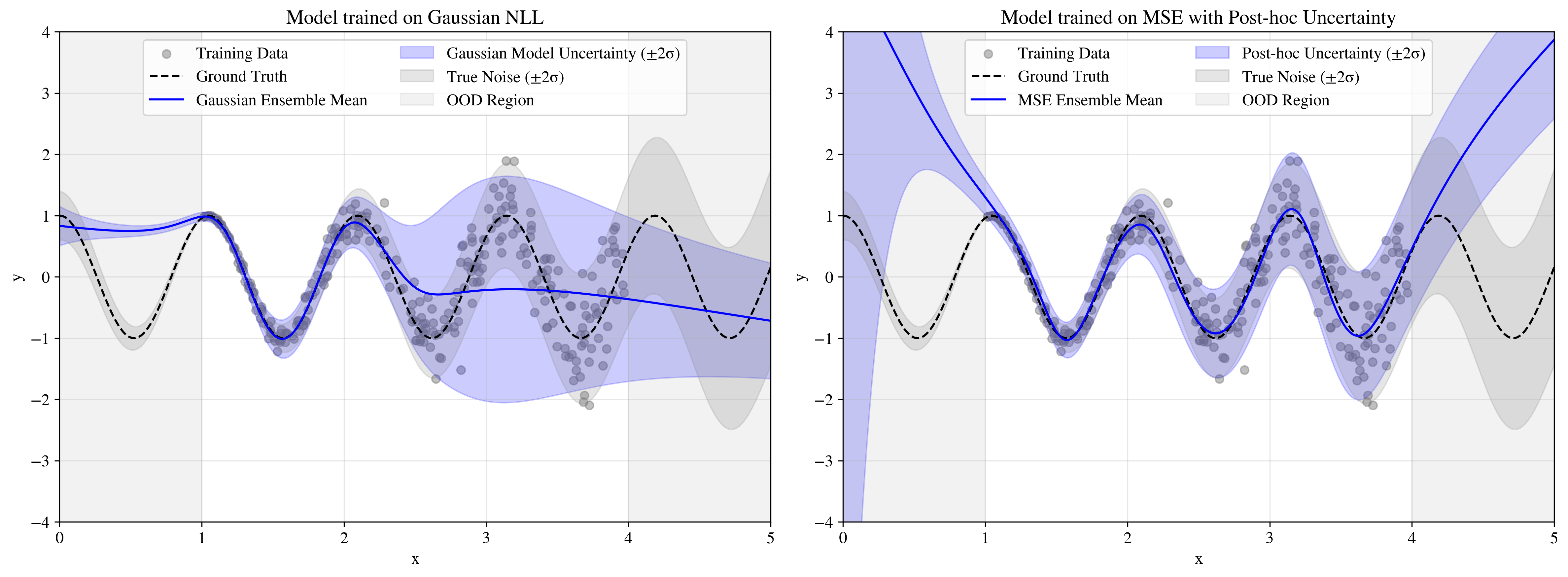}
    \caption{Comparison of (\textbf{left}) a model trained with two outputs and Gaussian NLL loss for 200 epochs and (\textbf{right}) a model trained with a single output and MSE loss, post-hoc uncertainty estimation via a probe dataset with the same number of epochs in total.}
    \label{fig:post-hoc_uq}
\end{figure}
\subsection{Approach}
Our post-hoc approach augments a frozen point estimator $f_{\theta}$ with an auxiliary variance head $g_{\phi}$. Trained on a small probe set, our method combines \emph{(i)} a detached Gaussian negative log-likelihood, \emph{(ii)} the original input $x$ to capture aleatoric noise, and \emph{(iii)} the frozen prediction $f(x)$ to capture relative epistemic effects. Here we outline justifications for each component before evaluating the framework.

\paragraph{(1) Detached Gaussian NLL objective.}\mbox{}\\
Given an MSE-trained predictor $f_{\theta}$, we optimize
\begin{equation}
    \phi^{\star}
    =\argmin_{\phi}\;\frac{1}{2N}\sum_{i=1}^{N}\frac{\bigl(\lfloor f(x_i;\theta)\rfloor-y_i\bigr)^{2}}{g_{\phi}(x_i,f(x_i))}+\frac{1}{2}\log g_{\phi}(x_i,f(x_i)).
    \label{eq:detached_nll}
\end{equation}
The gradient is \emph{detached} at $f(x_i)$, where $\lfloor \cdot \rfloor$ is the stop-gradient operator, so the base network remains untouched, thereby avoiding the instability that arises when mean and variance are trained jointly (cf.~\autoref{fig:post-hoc_uq}) and enabling post-hoc application. The objective is convex in $g_{\phi}$ and is optimized with AdamW.

\begin{figure}[t]
    \centering
    \includegraphics[width=\textwidth]{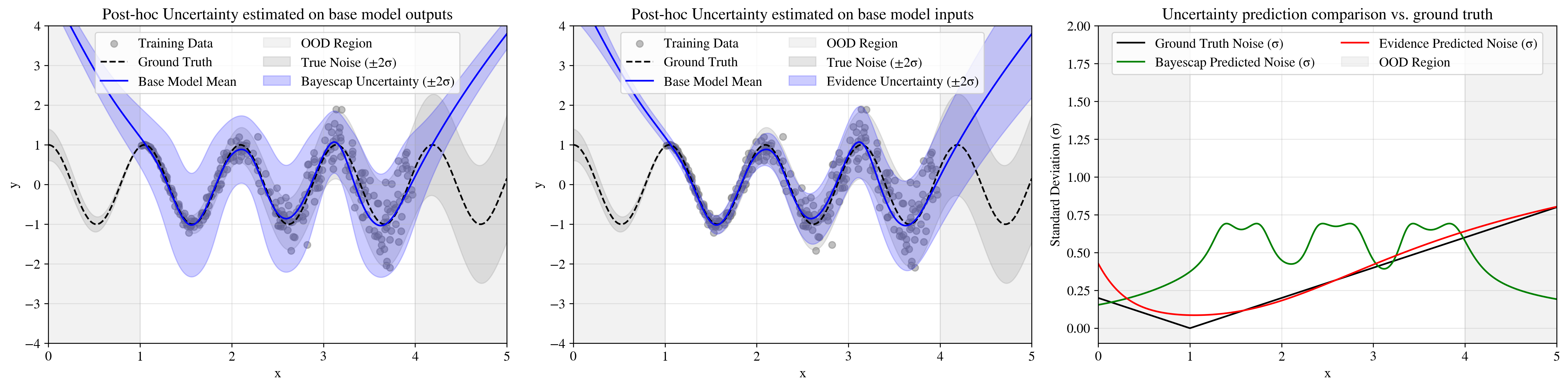}
    \caption{Post-hoc uncertainty estimation performance (\textbf{left}) when model is conditioned on the base model outputs $g(f(x))$ and (\textbf{center}) inputs $g(x)$. Plotting $g(f(x))$ against $g(x)$ (\textbf{right}) and the ground-truth noise shows that only the latter captures input-dependent uncertainty.}
    \label{fig:bayescap_evidence_comparison}
\end{figure}

\paragraph{(2) Conditioning on the input $x$.}\mbox{}\\
Access to the raw input is required to recover heteroscedastic noise $\sigma^{2}(x)=\operatorname{Var}(y\mid x)$. Without $x$, the estimator could at best learn a global scalar variance, i.e., the marginal variance $\mathrm{Var}(y)$. \autoref{fig:bayescap_evidence_comparison} illustrates the difference of conditioning on $x$ or $f(x)$ alone, with the right panel outlining that $x$ can be sufficient to capture aleatoric noise, while $f(x)$ alone leads to degenerate performance.

\paragraph{(3) Conditioning on the frozen prediction $f(x)$.}\mbox{}\\
The vector $f(x)$ can encode the base model's belief about its training manifold in dense output spaces. Its distance to that manifold correlates with epistemic risk and may be imbedded via latent structures, idiosyncratic to the mapping $f$. Augmenting $x$ with $f(x)$ therefore enables IO-CUE to inflate uncertainty on out-of-distribution inputs while preserving ID calibration by retrieving these idiosyncratic cues.

\textbf{Predictive uncertainty and model characterization.} When the frozen predictor fails, the detached NLL drives $g_{\phi}$ to inflate its variance. This behaviour can be formalized as follows.

Let $\mathcal{D} = \{(x_i, y_i)\}^n_{i=1}$ be i.i.d. samples from $y = \mu(x) + \epsilon, \quad \epsilon \sim \mathcal{N}(0, \sigma(x))$ with $p(x)$ supported on a compact set $\mathcal{S} \subset \mathcal{X}$. Let $f: \mathcal{X} \mapsto \mathbb{R}^K$ be a sufficiently expressive frozen regression model and $M = f(\mathcal{S})$ its compact image manifold. Finally, let $g: \mathcal{X} \times \mathbb{R}^K \mapsto \mathbb{R}^+$ be a universal function regressor. 

\begin{proposition}[Epistemic recovery]
There exist functions $g_a$ and $g_e$ such that
\[
    g_{a}(x)=\sigma(x),\quad g_{e}(f)=\lambda\,d\bigl(f,M\bigr),\;\lambda>0,
\]
where $d$ is a distance measure between predictions $f(x)$ and the support manifold $M$. Hence, $g_e$ is zero on in-distribution outputs and grows continuously off-manifold, providing a relative epistemic score.
\end{proposition}

\begin{corollary}
Minimizing \eqref{eq:detached_nll} on sufficiently diverse ID and OOD data learns $g_{\phi}\!\approx\!g_{a}+g_{e}$, i.e.
$\operatorname{Var}(y\mid x)\approx \sigma(x)+\lambda d(f(x),M).$ This will hold with $\lambda=1$ in all regions of $\mathcal{X}$ exposed to the post-hoc learner during training under reasonable assumptions of convergence.
\end{corollary}

\paragraph{IO--CUE.}\mbox{}\\
We call the resulting model, combining hybrid input-output conditioning and the detached NLL objective, \textcolor{RedViolet}{\textbf{I}nput--\textbf{O}utput-\textbf{C}onditioned \textbf{U}ncertainty \textbf{E}stimator (\textbf{IO--CUE})}:
\[
    \sigma_{\phi}(x,y)=g_{\phi}\bigl(x,\,f(x)\bigr).
\]

\paragraph{Paper road-map.} The next sections analyse IO--CUE from complementary angles:
\begin{itemize}[leftmargin=*]
    \item We demonstrate the effectiveness of this straightforward approach in general ID scenarios in \fullref{sec:uci_benchmarks} and \fullref{sec:depth_estimation}.
    \item \fullref{sec:ood_detection} examines the practical viablity of this approach. Here we will investigate how augmented data can be leveraged to improve OOD detection by learning inflated uncertainty forecasts from idiosyncratic input-output pairs.
    \item \fullref{sec:cross_network_generalization} ascertains that \textbf{IO-CUE} truly draws such information on predictvie uncertainty from model predictions $f(x)$ by testing its generalization on models trained on unseen datasets.
    \item \fullref{sec:increasing_input_perturbations} will further investigate the sensitivity of the proposed hybrid-input method to perturbations of the input space. This will confirm that \textbf{IO-CUE} continues to estimate an aleatoric component from inputs $x$.
\end{itemize}

\begin{figure}[ht]
    \centering
    \includegraphics[width=\textwidth]{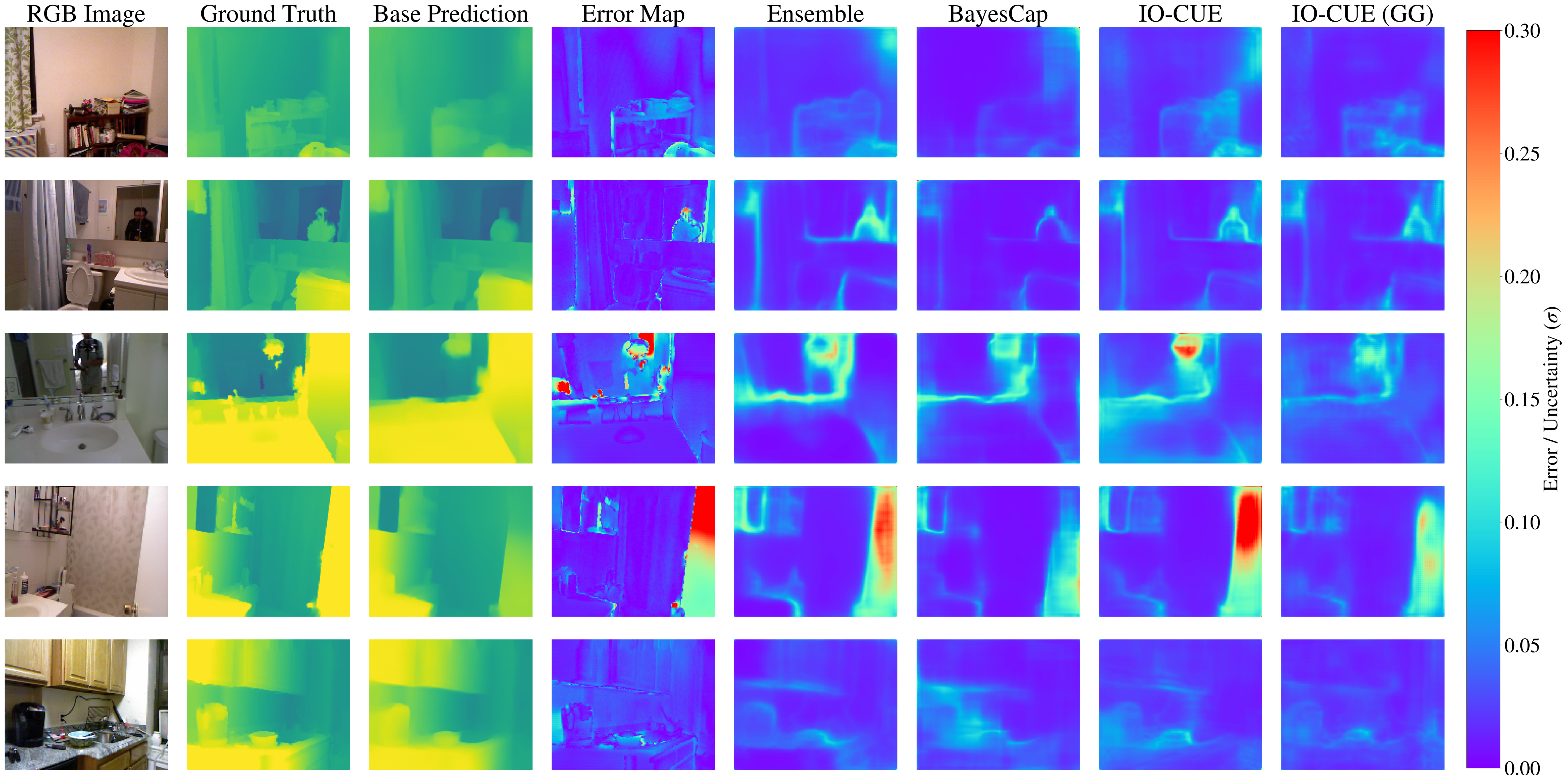}
    \caption{Qualitative comparison for depth estimation on the NYU Depth v2 test set. Results are comparable at a surface level. IO-CUE consistently achieves better NLL, ECE and EUC scores with results surpassing even the baseline Ensemble.}
    \label{fig:depth_uncertainty_comparison}
\end{figure}

\section{Experiments}
All experiments were conducted on a single NVIDIA H100 GPU. Runtimes ranged from a few minutes to about two hours for fitting the post-hoc models, where we also query predictions from the base model on each batch.

\subsection{Baselines}

\textbf{Traditional Ensemble} As a non-black-box baseline, we use a traditional ensemble of $N=5$ models where each model is trained on the NLL of the reported distribution, such that we can infer robust aleatoric and epistemic uncertainty \citep{lakshminarayanan2017simple,kendall2017uncertainties}. Generally, we aim to approach the performance of this baseline with our post-hoc methods.

\textbf{BayesCap} This approach is the closest in prior work and thus an indispensable member of our baseline suite. We train BayesCaps with a variety of distributions and training data sizes. The implementation is lifted directly from the publicly available code and modified only to fit our infrastructure \citep{upadhyay2022bayescap}.

\subsection{UCI Benchmarks}
\label{sec:uci_benchmarks}
\begin{table}[H]
    \caption{UCI Benchmark Results}
    \small
    \centering
    \begin{tabular}{l|ccc|ccc}
    \hline
    Dataset & \multicolumn{3}{c|}{NLL} & \multicolumn{3}{c}{ECE} \\
      & \makecell{Gaussian\\Ensemble} & BayesCap & IO-CUE & \makecell{Gaussian\\Ensemble} & BayesCap & IO-CUE \\
    \hline
    Boston & 4.58 $\pm$ 0.52 & 2.90 $\pm$ 0.26 & \textbf{2.44 $\pm$ 0.07} & 0.14 $\pm$ 0.02 & 0.08 $\pm$ 0.02 & \textbf{0.05 $\pm$ 0.05} \\
    Concrete & 4.01 $\pm$ 0.45 & 2.99 $\pm$ 0.20 & \textbf{2.85 $\pm$ 0.05} & 0.14 $\pm$ 0.02 & \textbf{0.03 $\pm$ 0.02} & 0.06 $\pm$ 0.01 \\
    Energy & 2.32 $\pm$ 0.57 & \textbf{1.31 $\pm$ 0.08} & 1.39 $\pm$ 0.11 & 0.17 $\pm$ 0.02 & 0.08 $\pm$ 0.02 & \textbf{0.02 $\pm$ 0.02} \\
    Kin8nm & -0.48 $\pm$ 0.13 & -0.88 $\pm$ 0.04 & \textbf{-0.90 $\pm$ 0.02} & 0.14 $\pm$ 0.00 & \textbf{0.13 $\pm$ 0.00} & \textbf{0.13 $\pm$ 0.00} \\
    Naval & -7.78 $\pm$ 0.06 & -7.11 $\pm$ 0.15 & \textbf{-7.78 $\pm$ 0.05} & \textbf{0.00 $\pm$ 0.01} & 0.14 $\pm$ 0.03 & 0.01 $\pm$ 0.01 \\
    Power & 2.69 $\pm$ 1.65 & 2.73 $\pm$ 0.00 & \textbf{2.73 $\pm$ 0.00} & 0.29 $\pm$ 0.01 & 0.02 $\pm$ 0.01 & \textbf{0.00 $\pm$ 0.00} \\
    Protein & 4.22 $\pm$ 0.34 & 2.97 $\pm$ 0.12 & \textbf{2.68 $\pm$ 0.02} & 0.13 $\pm$ 0.00 & 0.05 $\pm$ 0.00 & \textbf{0.00 $\pm$ 0.00} \\
    Wine & 3.11 $\pm$ 0.00 & 1.22 $\pm$ 0.00 & \textbf{0.92 $\pm$ 0.00} & 0.17 $\pm$ 0.00 & 0.12 $\pm$ 0.00 & \textbf{0.03 $\pm$ 0.00} \\
    Yacht & 0.64 $\pm$ 0.00 & \textbf{-2.24 $\pm$ 0.00} & 1.17 $\pm$ 0.00 & 0.08 $\pm$ 0.00 & \textbf{0.02 $\pm$ 0.00} & 0.24 $\pm$ 0.00 \\
    \hline
    \end{tabular}
    \vspace{1em}
    \label{tab:uci_benchmarks}
\end{table}
As a sanity check on real-world datasets, we report results for a suite of commonly used UCI regression benchmarks \citep{hernandez2015probabilistic,gal2016dropout,amini2020deep}. We report NLL and expected calibration error (ECE) \citep{kuleshov2018accurate} for each model including a probabilistic ensemble baseline. The base model used for post-hoc training and the Gaussian ensemble baseline use $N=5$ members and consist of three layers with 100 units each. The post-hoc models (BayesCap and IO-CUE) comprise three layers with 50 units each (comprehensive training details are provided in \autoref{sec:training_details}). Results in \autoref{tab:uci_benchmarks} show that both post-hoc methods significantly outperform the probabilistic ensemble baseline with concurrent parameter estimation. We find that IO-CUE exceeds the performance of BayesCap with few exceptions. The nonetheless competitive performance of BayesCap, however, suggests that even in this simple setting, with a single target variable, the underlying base model exposes significant information about aleatoric uncertainty.

\subsection{Depth Estimation}
\label{sec:depth_estimation}
For depth estimation, we train a U-Net style architecture \citep{falk2019u} on the NYU Depth V2 dataset \citep{silberman2012indoor} as seen in \citep{amini2020deep}, i.e., a variant with random crops amounting to ~27k total samples. We use the same train/test-splits as reported there. In addition to the mean square error, we impose an image gradient based penalty \citep{hu2019revisiting} to enfore clean edges in the depth predictions. Where necessary, the model will have multiple heads to output disjoint parameter maps. Each head consists of a final three convolutional layers. The same is true for post-hoc models which estimate multiple parameters. Post-hoc models use the same U-Net architecture as the base model (the impact of model size is examined in \autoref{sec:model_size}) and are trained on a probe dataset consisting of 10\% or $\sim3000$ samples of the original training data at a batch size of $32$ and for $30.000$ iterations (we investigate the impact of probe dataset size in \autoref{sec:probe_dataset_size}). We report ECE, NLL, and error-uncertainty correlation (EUC) (Spearman's rank) for each model including ensemble baselines across 8 runs. In addition to a Gaussian IO-CUE, we provide results for a Generalized Gaussian-based IO-CUE (GG).

\begin{wraptable}{r}{0.6\textwidth}
    \caption{NYU Depth v2 Dataset Results}
    \small
    \centering
    \hspace*{-.75\columnsep}\begin{tabular}{l|c|c|c}
    \hline
    Method & ECE & NLL & EUC \\
    \hline
    Ensemble & 0.01 $\pm$ 0.00& -1.72 $\pm$ 0.07 & 0.40 $\pm$ 0.01 \\
    BayesCap & 0.12 $\pm$ 0.02 & -1.17 $\pm$ 0.02 & 0.35 $\pm$ 0.01 \\
    IO-CUE & \textbf{0.00 $\pm$ 0.00} & \textbf{-2.33 $\pm$ 0.04} & \textbf{0.53 $\pm$ 0.03} \\
    IO-CUE (GG) & \textbf{0.00 $\pm$ 0.00} & -2.15 $\pm$ 0.01 & 0.47 $\pm$ 0.02 \\
    \hline
    \end{tabular}
    \label{tab:nyu_depth_results}
\end{wraptable}

We find that IO-CUE manages to dominate across all metrics \autoref{tab:nyu_depth_results}. This suggests that our method's uncertainty forecasts are more informative for identifying erroneous predictions (higher EUC: $0.53$ vs. $0.35$), are statistically consistent with true model failures (lower ECE: $<0.0001$ vs. $0.15$), and sharper than BayesCap's estimates (lower NLL: $-2.33$ vs. $-1.17$). Since all post-hoc models are trained on an MSE-optimized base, RMSE is lower than in the Gaussian Ensemble baseline. As suggested in the introductory sections, this highlights one of the major benefits of post-hoc uncertainty estimation. Namely the non-existant impact on base model performance often exhibited in concurrent uncertainty estimation. We provide qualitative results in \autoref{fig:depth_uncertainty_comparison}.

\section{Analysis}
\begin{figure}[t]
    \centering
    \begin{subfigure}[b]{\textwidth}
        \centering
        \includegraphics[width=\textwidth]{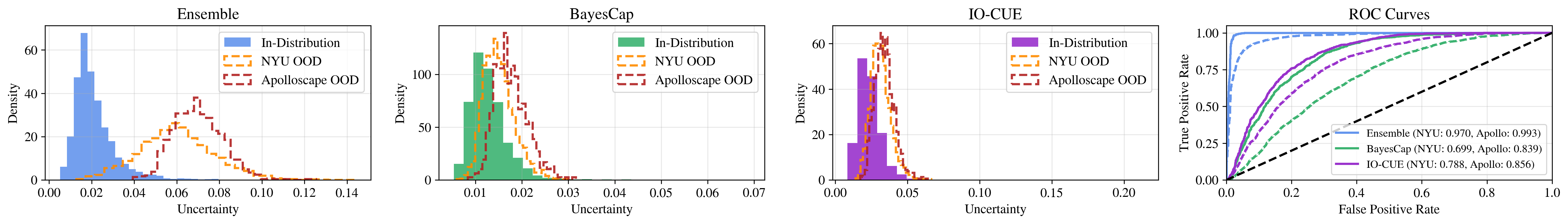}
        \label{fig:ood_nyu}
    \end{subfigure}
    \vspace{-1em}
    \begin{subfigure}[b]{\textwidth}
        \centering
        \includegraphics[width=\textwidth]{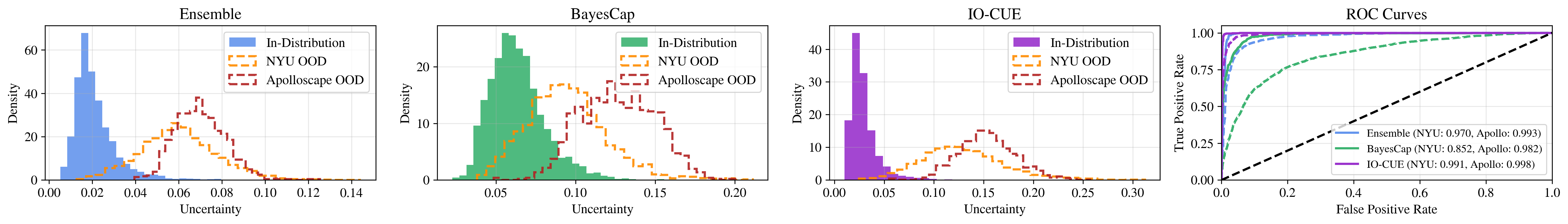}
        \label{fig:ood_apollo}
    \end{subfigure}
    \caption{OOD Detection Results before (\textbf{top}) and after (\textbf{bottom}) data augmentation for a flipped version of the NYU Depth v2 dataset and the previously unseen Apolloscape dataset. Augmentation improves model characterization by post-hoc uncertainty quantifiers dramatically. (ROC curves: Solid lines: Apolloscape, Dashed lines: flipped NYU.)}
    \label{fig:ood_analysis}
\end{figure}

\subsection{Post-Hoc OOD Detection and Model Characterization}
\label{sec:ood_detection}
Detecting OOD samples in a post-hoc manner is particularly challenging under black-box constraints, where the variance model has no access to the base model’s parameters or uncertainty proxies like ensemble disagreement. We evaluate this in the context of depth estimation, where the output space is structured and potentially informative of epistemic uncertainty.

We first test all models on a horizontally flipped version of the NYU Depth v2 dataset to simulate near-distribution shifts, limiting the utility of low-level image statistics. This setup ensures that base model errors predominantly stem from epistemic uncertainty. Subsequently, we evaluate on the fully OOD ApolloScape dataset \citep{huang2018apolloscape}, which introduces both covariate and semantic shifts.

Results indicate that in the near-distribution setting, both post-hoc models underestimate predictive error (\autoref{fig:ood_analysis}, top), though IO-CUE yields better AUROC scores than BayesCap (0.79 vs. 0.70), while the ensemble remains significantly stronger (0.97). Similar trends hold on ApolloScape, where performance of post-hoc models improves slightly but remains far below the ensemble. These outcomes reflect the limitations of black-box post-hoc approaches.

To address this, we augment the probe dataset with random Gaussianblur and Colorjitter during post-hoc training, augmentations notably disjoint of horizontal flios. This simple intervention increases the diversity of base model behaviors observed by the uncertainty estimator. When re-evaluated, both post-hoc models show marked improvement on both OOD datasets (\autoref{fig:ood_analysis}, bottom), validating the hypothesis that a wider range of model idiosyncrasies enhances generalization to OOD settings and yields more informative predictive uncertainty.

\subsection{Cross-network generalization}
\label{sec:cross_network_generalization}
Effective OOD detection on unseen samples suggests that post-hoc models conflate aleatoric forecasts with a type of model uncertainty which is derived from highly structured base model outputs. To test this hypothesis, we train four separate base models on NYU Depth v2 with different augmentations (Flip ($p=0.5$), Colorjitter (brightness, contrast, saturation, hue$=0.1$), Gaussianblur ($\sigma \in \{0.01, 1.0\}$), Grayscale ($p=0.5$)). During testing, all $p=1.0$ and $\sigma=1.0$ is fixed. We again employ the previously trained IO-CUE model, which has only seen unaugmented samples from the same dataset. To establish reliance on structures in model outputs, we confront the post-hoc model with counterfactual inputs, i.e., uncorrupted inputs but corrupted model outputs $g(x, f(x_{\sigma}))$. We then perform the OOD detection task setting each base model against each augmented dataset using only the previously trained IO-CUE model.
\begin{figure}[h]
    \centering
    \includegraphics[width=\textwidth]{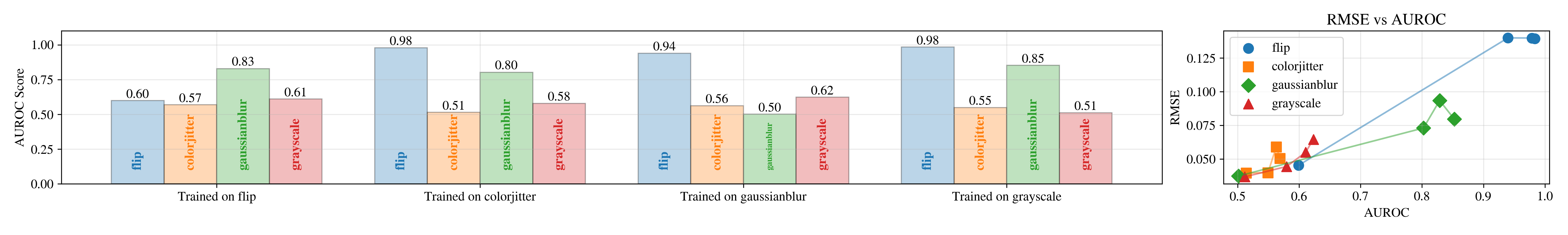}
    \caption{AUROC scores for multiple base models on differently augmented datasets in a counterfactual setting $g(x, f(x_{\sigma}))$ (\textbf{left}). Our approach manages to predict uncertainties related to the true model performance, suggesting that markers of uncertainty can be extracted from the base model outputs. Plotting AUROC against model error reveals the general trend (\textbf{right}).}
    \label{fig:ood_cross_network_generalization}
\end{figure}
For all four base models, IO-CUE-based AUROC is lowest when the model is set against a known augmentation, suggesting that the post-hoc model does not estimate higher uncertainties for that specific dataset, despite being unfamiliar with the augmentation in question \autoref{fig:ood_cross_network_generalization} (left). The effect is most pronounced in the Gaussianblur and Flip scenarios where uncertainties are highly predictive of model error. The applied augmentations do not have well-seperated effects. For example, a model trained with Colorjitter has lower error on the Grayscale dataset because of some lateral generalization. While the post-hoc estimator fails to generalize absolute error forecasts, it predicts informative relative uncertainties across all scenarios, suggesting that it captures epistemic uncertainty conditioned on $f(x)$. In fact, AUROC increases as model error increases as a general trend \autoref{fig:ood_cross_network_generalization} (right). 

\begin{figure}[H]
    \centering
    \includegraphics[width=\textwidth]{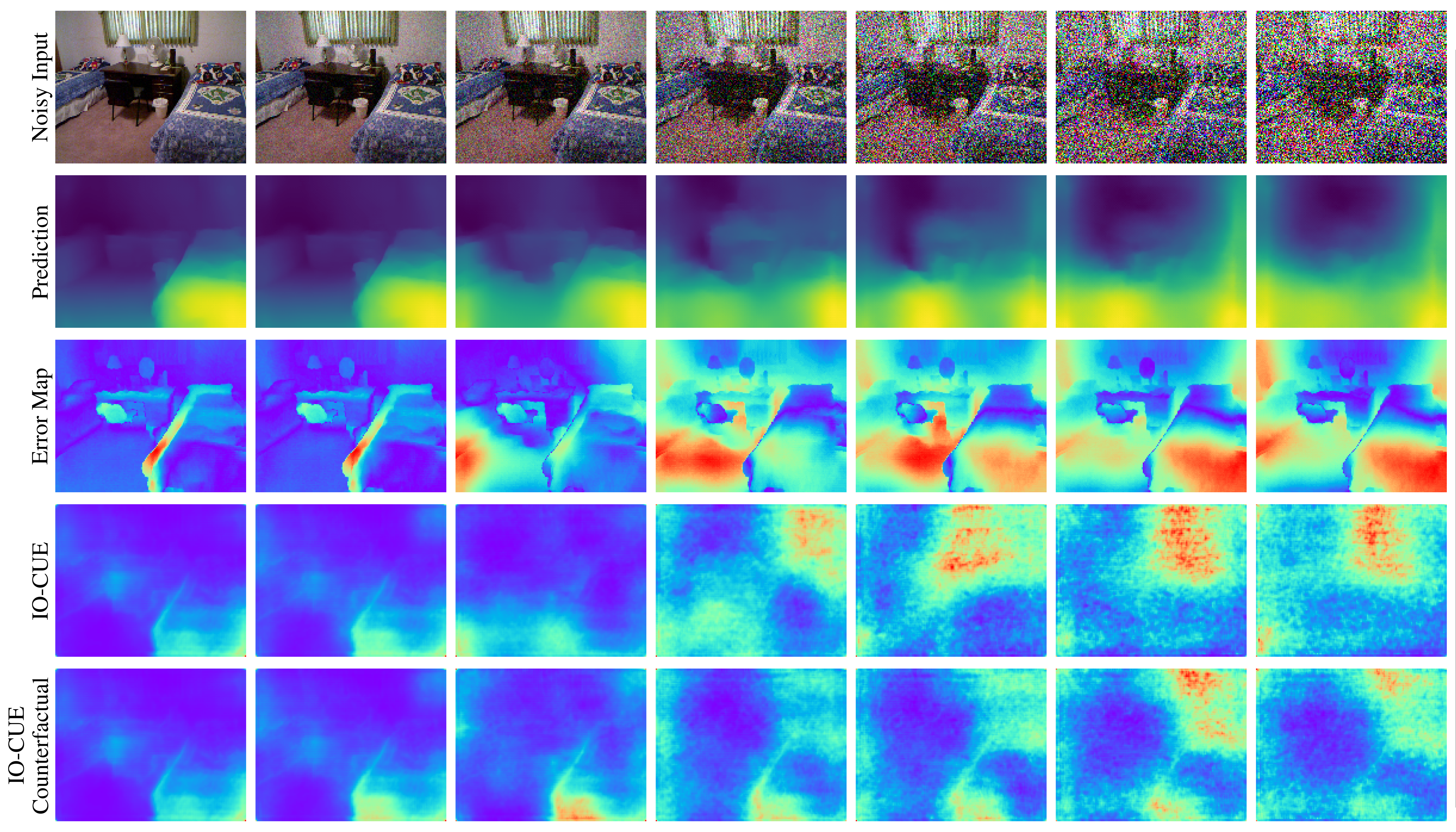}
    \caption{Qualitative examples of the effect of input perturbations on the uncertainty estimates of our approach. The trend of increasing uncertainty estimates holds even in the ablation condition, where $\sigma = g(x_{\sigma}, f(x))$, suggesting that the post-hoc estimator is sensitive to the input and does not entirely rely on the base model outputs.}
    \label{fig:perturbation_examples}
\end{figure}

\subsection{Increasing input perturbations}
\label{sec:increasing_input_perturbations}
\begin{wrapfigure}{r}{0.5\textwidth}
    \vspace{-1.85em}
    \centering
    \begin{minipage}{0.48\linewidth}
        \includegraphics[width=\linewidth]{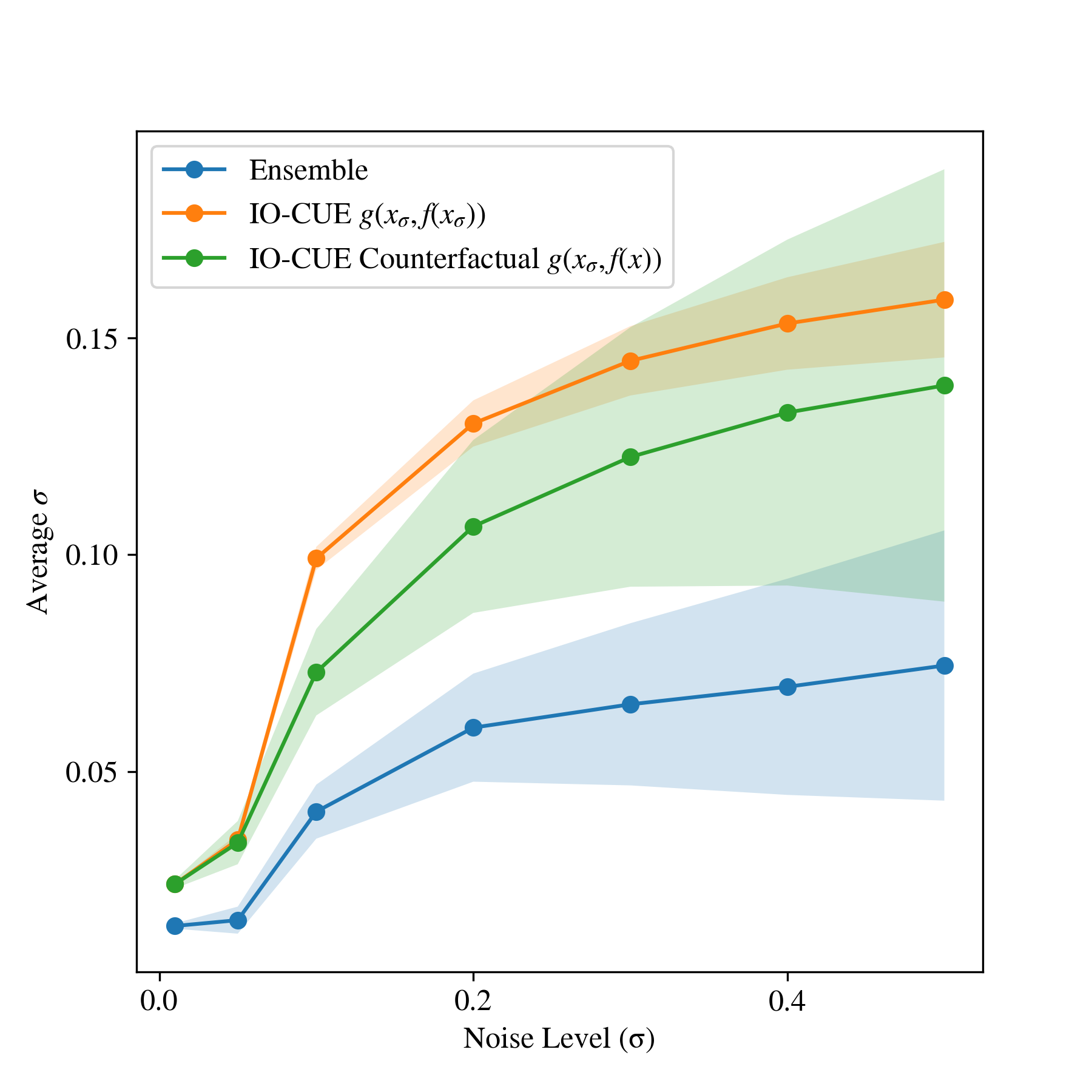}
    \end{minipage}
    \hfill
    \begin{minipage}{0.48\linewidth}
        \includegraphics[width=\linewidth]{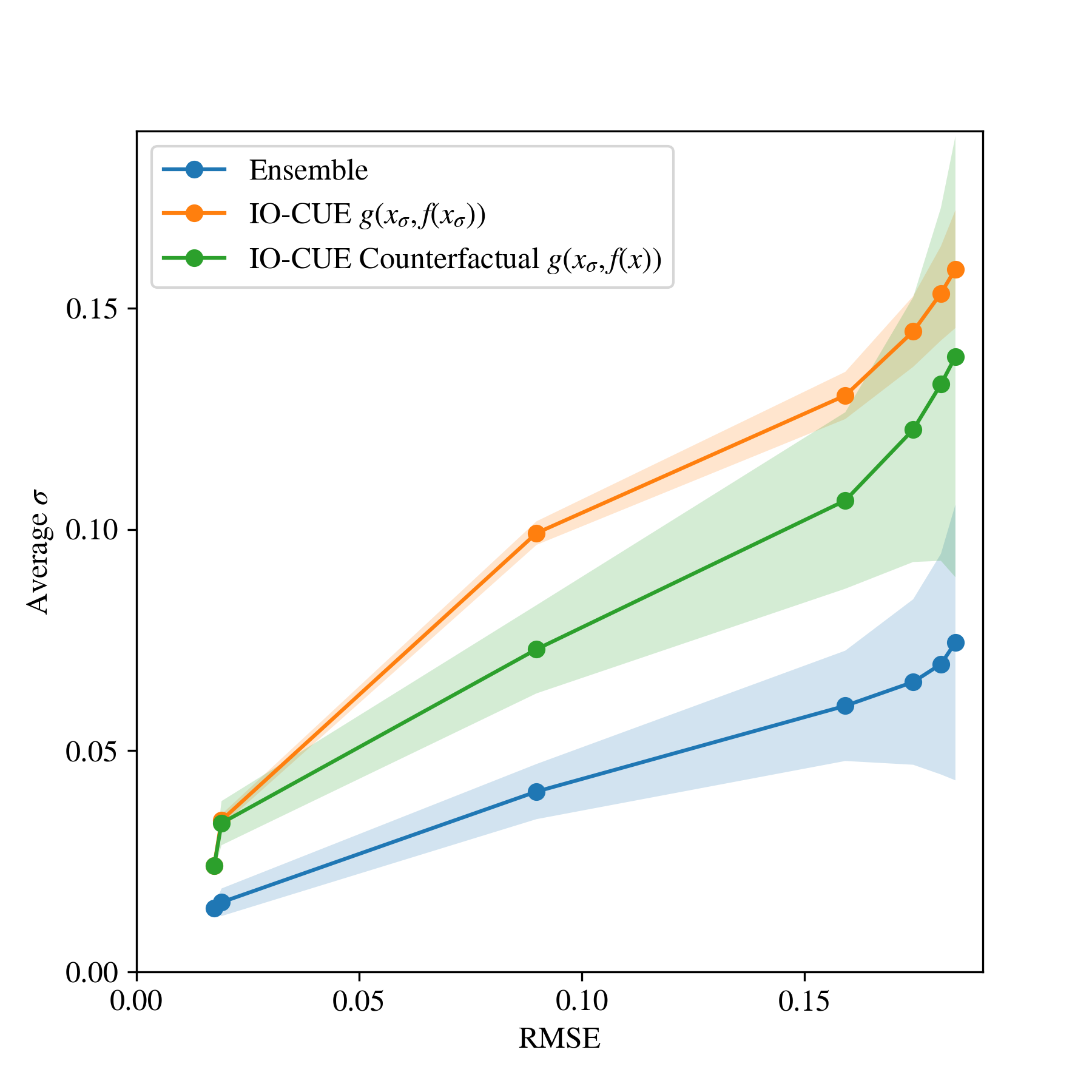}
    \end{minipage}
    \caption{Results of increasing input perturbations. (\textbf{Left}) Average uncertainty scores w.r.t. noise levels. (\textbf{Right}) Uncertainty forecasts vs. base model RMSE.}
    \label{fig:increasing_input_perturbations}
\end{wrapfigure}
We need to confirm that IO-CUE does not rely on $f(x)$ alone to infer uncertainties. This would preclude the detection of true aleatoric, i.e., data-dependent uncertainty in favor of forecasts compounded with model-dependent effects. As such, we employ a canonical experiment involving seven increasing levels of input sample corruption:  $x_{\sigma} = \textrm{clamp}(x + \epsilon, 0, 1) \textrm{ and } \epsilon \sim \mathcal{N}(0, \sigma), \quad \forall \sigma \in [0.01, 0.05, 0.1, 0.2, 0.3, 0.4, 0.5,].$

We then analyse average uncertainty scores w.r.t. noise levels and compare uncertainty forecasts with base model RMSE. As before, we employ a counterfactual scenario. Here, we highlight the sensitivity to $x$ by comparing $g(x_{\sigma}, f(x_{\sigma}))$ and $g(x_{\sigma}, f(x))$. The difference between these estimates serves as a proxy of the sensitivity to $x$ and $f(x)$. \footnote{We do not test BayesCap in this scenario, since its formulation $g(f(x))$ makes it entirely dependent on base model outputs.} Results indicate that our approach is indeed sensitive to variations of the input sample, with increasing uncertainty estimates as noise levels increase \autoref{fig:increasing_input_perturbations} (left). The effect is marginally less pronounced in the ablation condition, where $\sigma = g(x_{\sigma}, f(x))$ but the general trend is largely preserved across all noise levels confirming appropriate attribution of aleatoric uncertainty. Qualitative examples of the effect are shown in \autoref{fig:perturbation_examples}.

\section{Conclusion}
We have presented a principled framework for post-hoc uncertainty quantification grounded in maximum likelihood estimation and sequential optimization. Building on prior work, we formalized the necessary conditions under which variance can be reliably estimated from frozen model outputs, clarifying longstanding assumptions with theoretical rigor. While earlier approaches have employed $f(x)$ for post-hoc UQ, they lacked justification and failed to assess its expressiveness and limitations. Our analysis establishes when and why $f(x)$ may encode quasi-epistemic signals. Taken together, our IO-CUE method constitutes a definitive formulation of optimization-based post-hoc UQ and a promising starting point for applications in other problems.

\subsection*{Limitations}
\label{sec:limitations}
In this work, we provide a principled approach to post-hoc uncertainty quantification. However, these principles are, for now, limited to regression problems. Furthermore, the results presented make the Gaussian assumption, which requires that the frozen base models are found via MSE methods. Only this guarantees that they will coincide with the MLE of the mean. While manipulations of the MSE, such as via image-gradient penalties, have no adverse effects in our experiments, the applicability of our method has to be investigated on further distribution families and problems in future work. Regardless, its principles are firmly grounded in established research and offer a jumping-off point for further investigations.

\subsubsection*{Acknowledgements}
We thank Seong Joon Oh for supporting this project with his expertise and insights.

This research was funded by the BMFTR project 'HoLoDEC' (grant 16ME0695K-16ME0705), the European Regional Development Fund (ERDF) within the project 'AIDA' (grant FEIH 2476184) and the Horizon Europe research and innovation programme under project 'HEIDI' (Grant Agreement No. 101069538). The authors would like to thank the consortium for the successful cooperation. Views and opinions expressed are those of the authors only and do not necessarily reflect those of the European Union. Neither the European Union nor the granting authority can be held responsible for them.

\bibliography{proposal.bib}
\bibliographystyle{plainnat}

\newpage

\section*{Supplementary Materials}
\appendix

\section{Training Details}
\label{sec:training_details}
\subsection{UCI benchmark}
All models were trained with the same regime, i.e., hyperparameters were not optimized per individual dataset. However, best hyperparameters for the learners were chosen based on performance on the \emph{Boston Housing}\footnote{The authors acknowledge the divisive nature of this benchmark. We make no statement on its biases or assumptions and simply present results for ease of comparison.} benchmark set. As a result, we equip the base models with three layers of 100 units each, with ReLU activations before the ultimate layer. We use the Adam optimizer with a learning rate of 0.0001 for the base models. We set the number of epochs to 100, with an option for early stopping if 8 consecutive evaluations after each epoch have not resulted in improved NLL. In such a case, we load the last best performing model and terminate the training.

Each post-hoc model was likewise comprised of three layers, however, each layer held 50 units. Both models were found to perform best with intermittent Tanh activations. They were trained with the AdamW optimizer with a learning rate of 0.0001 and weight decay at 0.01. The maximum number of epochs was set to 100, with the same early stopping procedure as the base models. The batch size was 128 for all models. 

Each dataset was subjected to standard scaling during learning, but all predictions and ground-truth values returned to original scale for evaluating relevant metrics.

\subsection{Depth estimation benchmark}
Training on the NYU Depth v2 dataset was performed on the entirety of the samples for the base models, such as the dataset appears in \citep{amini2020deep}, i.e., with additional random crops and a total of about 27k samples. Post-hoc models were unless otherwise stated trained on 10\% of this data. All models were trained at a batch size of 32 and an image size of (160, 128) and no augmentations, again, unless stated otherwise.

Base models were trained with the Adam optimizer, at a learning rate of 0.00005, and cosine annealing over a fixed 64 epochs. Instead of early stopping, we chose the best performing model after training by investigating MLE and NLL scores. Dropout rates were fixed to $p=0.2$.

Post-hoc models were instead trained on a fixed number of update steps, in order to account for roughly 1/2 of the total training duration of the base model. Image inputs were of the same size, however, they were concatenated with the depth prediction of the underlying base model. We used the Adam optimizer with same learning rate here, as in the base models. During training, we set the dropout probability to $p=0.3$.

\section{Impact of model size}
\label{sec:model_size}
\begin{wraptable}{r}{0.5\textwidth}
    \caption{Model size configurations.}
    \small
    \centering
    \hspace*{-.75\columnsep}
    \centering
    \begin{tabular}{lccc}
        \toprule
        Model & Encoder & Decoder & Max \\
              & Blocks  & Blocks  & Channels \\
        \midrule
        Small        & 3 & 2 & 128 \\
        Medium       & 4 & 3 & 192 \\
        Large (default) & 5 & 4 & 512 \\
        \bottomrule
    \end{tabular}
    \label{tab:model_sizes}
\end{wraptable}

Here, we investigate the impact of model complexity on the performance of IO-CUE. Our default model, used throughout the main text, is a U-Net encoder-decoder architecture with 5 encoder blocks consisting of 2 convolutional layers, followed by max-pooling and 4 decoder blocks that constist of one ConvTranspose layer, each followed by 2 convolutional layers. The final output is generated via a 3-layer convolutional head. We compare the performance of this model with a small and medium sized variant \autoref{tab:model_sizes}.

\begin{figure}[H]
    \centering
    \includegraphics[width=\textwidth]{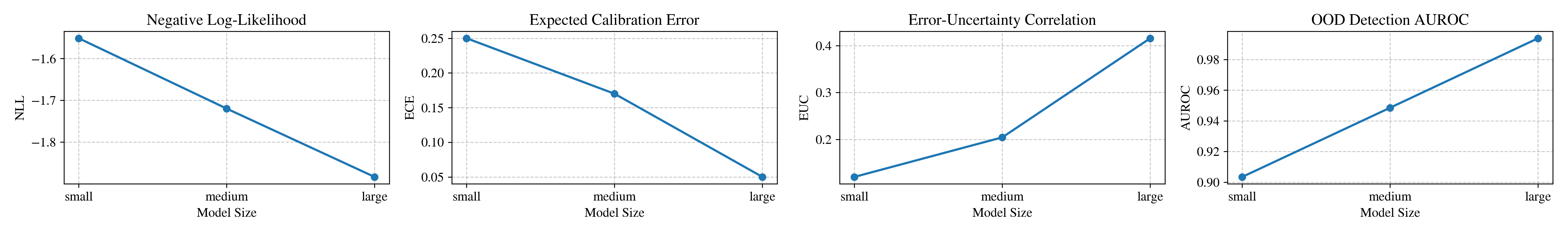}
    \caption{Impact of model size on IO-CUE performance.}
    \label{fig:model_size}
\end{figure}

Each model was trained for $10.000$ steps, using the same training regime as before. Additionally, we applied the flip transform to improve OOD detection performance. OOD detection was performed on the ApolloScape dataset. As expected, metrics noticably improve with model size. We observe generally that a model of comparable complexity to the base model is required in order to recover canonical errors with some reliability. Using the flip transform with $p=0.5$ improved OOD detection across all models dramatically (from AUROC 0.5 to 0.9 for the small and 0.6 to 0.99 for the large variant) \autoref{fig:model_size}.

\section{Impact of probe dataset size}
\label{sec:probe_dataset_size}
In another experiment, we investigate the impact of the probe dataset size on the performance of IO-CUE. We use the same large default model and train on 10\%, 50\%, and 100\% of the NYU Depth v2 dataset. Beyond this, we apply the same training regime, including flip transform with $p=0.5$, and a total of $10.000$ training steps. As such, each model receives the same number of gradient updates, with probe dataset size accounting for diversity and richness alone.

\begin{figure}[H]
    \centering
    \includegraphics[width=\textwidth]{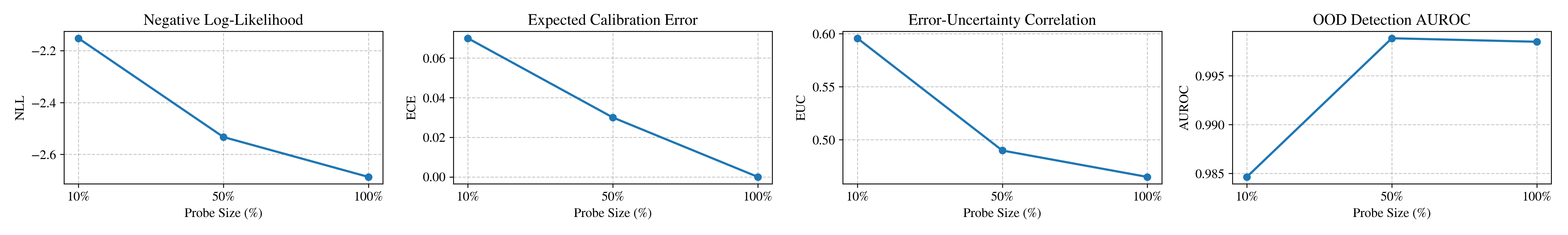}
    \caption{Impact of probe dataset size on IO-CUE performance.}
    \label{fig:probe_dataset_size}
\end{figure}

Comparing the size of probe datasets, we observe a curious effect where increasing probe sizes lead to improved NLL and ECE, but decreasing EUC. Since all models are equal and trained with the same regime, we can discount explanations like over-regularization or impacts of architectural biases. It is clear that training on smaller probe sizes leads to effectively ranked uncertainty estimates that are not precise estimates of density. This means that the model does know when it is wrong, but not necessarily by how much. Even at very small probe sizes, IO-CUE is effective at discriminating between ID and OOD samples, provided some augmentations are applied to the probe dataset.

\end{document}